\documentclass[9pt,journal]{IEEEtran}  

\IEEEoverridecommandlockouts                              

\usepackage{graphics} 
\usepackage{epsfig} 
\usepackage{mathptmx} 
\usepackage{times} 
\usepackage{amsmath} 
\usepackage{amssymb}  
\usepackage{url}
\usepackage{romannum}
\usepackage{xcolor}
\usepackage{float}
\usepackage{url}
\usepackage[ruled,vlined]{algorithm2e}
\usepackage{array}
\usepackage{placeins}

\title{\LARGE \bf
ConvSequential-SLAM: A Sequence-based, Training-less Visual Place Recognition Technique for Changing Environments 
}

\author{Mihnea-Alexandru Tomiță$^{1}$, Mubariz Zaffar$^{1}$, Michael Milford$^{2}$, Klaus McDonald-Maier$^{1}$ and Shoaib Ehsan$^{1}$
\thanks{$^{1}$Authors are with the School of Computer Science and Electronic Engineering,
        University of Essex, CO4 3SQ, United Kingdom
        {\tt\small matomi@essex.ac.uk, mubariz.zaffar@essex.ac.uk,  kdm@essex.ac.uk, sehsan@essex.ac.uk}. }%
\thanks{$^{2}$Michael Milford is with the School of Electrical Engineering and Computer Science, Queensland University of Technology, Brisbane, QLD 4000, Australia
        {\tt\small michael.milford@qut.edu.au}}%
        
\thanks{This work is supported by the UK Engineering and Physical Sciences Research Council through grants EP/R02572X/1 and EP/P017487/1.}
}

\begin{document}

    \maketitle
    \thispagestyle{empty}
    \pagestyle{empty}

    \begin{abstract}
        Visual Place Recognition (VPR) is the ability to correctly recall a previously visited place under changing viewpoints and appearances. A large number of handcrafted and deep-learning-based VPR techniques exist, where the former suffer from appearance changes and the latter have significant computational needs. In this paper, we present a new handcrafted VPR technique that achieves state-of-the-art place matching performance under challenging conditions. Our technique combines the best of 2 existing trainingless VPR techniques, SeqSLAM and CoHOG, which are each robust to conditional and viewpoint changes, respectively. This blend, namely ConvSequential-SLAM, utilises sequential information and block-normalisation to handle appearance changes, while using regional-convolutional matching to achieve viewpoint-invariance. We analyse content-overlap in-between query frames to find a minimum sequence length, while also re-using the image entropy information for environment-based sequence length tuning. State-of-the-art performance is reported in contrast to 8 contemporary VPR techniques on 4 public datasets. Qualitative insights and an ablation study on sequence length are also provided.  
        
        \endgraf
        
    \end{abstract}
    

    \section{Introduction}
        To autonomously operate in an environment, a mobile robot has to map, localise and navigate through the environment. This problem of simultaneously mapping and localising the environment is a widely researched topic within autonomous robotics, termed as Simultaneous Localisation and Mapping (SLAM) \cite{cadena2016past}. Generally, robots are equipped with a wide variety of sensors such as cameras, lasers and wheel encoders, that 
        provide essential information that enables motion and location estimates. However, iterative location estimates based on dead-reckoning accumulate errors, which become significant over longer trajectories, leading to incorrect belief about the robot's location in the world. Within autonomous robotics, these accumulated errors can be catered-for, if the robot revisits and recognises a previously visited (known) place in the world, which is generally labelled as `loop-closure'. For  a vision-only system, this loop-closure can be achieved if a robot is able to recall a previously visited place using only visual information. This task has become a subject of great interest within the robotic vision community and therefore Visual Place Recognition (VPR) has developed as a dedicated field within autonomous robotics over the past 15 years \cite{VPRsurvey}. 
        
\begin{figure}[t]
        \begin{center}
        \includegraphics[width=1\linewidth]{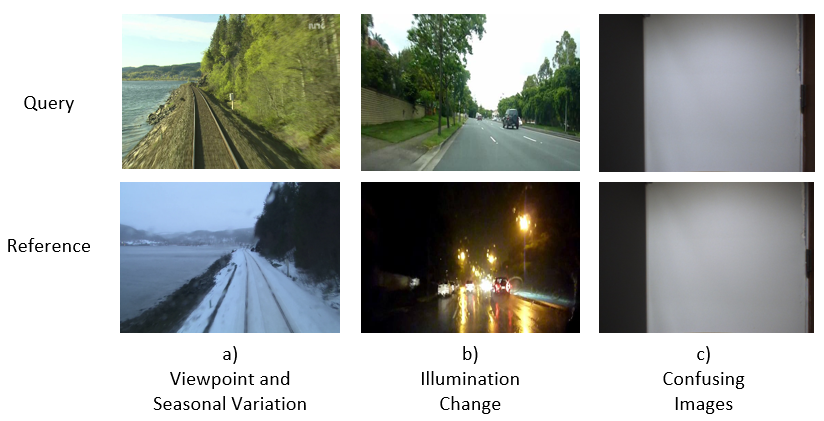}   
        \end{center} 
        \caption{Images showing the same place under different conditions, a) Different viewing angle and seasonal change (summer to winter), b) Illumination variation (day to night), c) Confusing and feature-less images that can lead to perceptual aliasing.}
        \label{vprchallenges}
        \vspace{-5mm}
        \end{figure}

        VPR, is significantly challenging due to the variations in viewpoint, illumination, seasons and dynamic objects, as depicted in Fig. \ref{vprchallenges}. Due to these variations in the environment, the appearance of a given place can change in a drastic manner between repeated traversals, thus making it challenging for a VPR system. Moreover, confusing and feature-less frames can also drastically increase the difficulty in place matching \cite{zaffar2018memorable} due to the lack of distinct features to distinguish a given place from a geographically-different but visually-similar place (perceptual-aliasing). To achieve loop-closure in a SLAM system or localisation in a fixed-size, visually-mapped environment, each query image (camera frame) has to be matched with the appropriate reference image from the map, taking into consideration all the major changes that a place can undergo given computational constraints. \endgraf 
         
        VPR is usually cast as an image retrieval problem. Prior to the usage of deep-learning in VPR systems, handcrafted local and global feature descriptors were used to perform place recognition. Local feature descriptors only process 
        salient parts (keypoints) of the image, while global feature descriptors process the entire image regardless of its content. The performance of local feature descriptors suffers under illumination changes in the environment, while global feature descriptors cannot handle viewpoint variation \cite{VPRsurvey}. The application of deep-learning, especially Convolutional Neural Networks (CNNs), was first studied by Chen \textit{et al.} in \cite{chen2014convolutional} and since then most of the advances in VPR have been primarily due to deep-learning-based techniques, as reviewed in Section \ref{literature_review}. CNNs are systems capable of learning features extracted from images using supervised training on labeled datasets. Such CNN-based VPR techniques have achieved state-of-the-art performance on the most challenging VPR datasets, as evaluated in \cite{zaffar2019levelling} and \cite{zaffar2019state}. However, in order to train a CNN for VPR tasks, one needs a large-scale dataset of labeled images taken from different environments, under various angles, seasons and illumination conditions. Although labelled VPR datasets exist, such as Oxford Robot Car dataset \cite{maddern20171}, SPED dataset \cite{chen2017deep} and Pittsburgh dataset \cite{torii2013visual}, they represent a particular environment under limited conditional and viewpoint changes. Therefore, the creation of a large-scale, labeled dataset representing all the different possible variations is not feasible and requires significant time and resources. Furthermore, training a CNN within reasonable times to adjust to a new environment will require dedicated Graphics Processing Units (GPUs) and may take several days/weeks in order to be trained. Because of the CNN's intense computational nature, their encoding-time and run-time memory are also significantly higher than those needed for the handcrafted feature descriptors. Although the CNN-based VPR techniques have largely outperformed handcrafted feature descriptor-based techniques on the image matching front, their intense computational requirements make them harder to use in this field. As a result of these demanding requirements, the deployment of CNN-based techniques for VPR are restricted for resource-constrained vehicles such as battery-powered aerial, micro-aerial and ground vehicles, as discussed in \cite{zaffar2019state} and \cite{maffra2019real}. 
        
        Contrary to the above understanding of the success of deep-learning for robustness to appearance changes, it has been previously shown in SeqSLAM \cite{milford2012seqslam} that simplistic handcrafted techniques can still be used to achieve robustness to conditional variations by employing sequential/temporal information (i.e, consecutive image frames). However, SeqSLAM cannot handle viewpoint variations and its invariance to illumination changes can be improved by replacing simplistic pixel matching with illumination-invariant descriptors. Moreover, the fixed and predefined sequence length in SeqSLAM renders it inflexible to different environments. More recently in CoHOG \cite{zaffar2020cohog}, a fully handcrafted and training-less VPR technique, it was shown that state-of-the-art matching performance can be achieved by handcrafted techniques on viewpoint-variant datasets. Although CoHOG was shown to be viewpoint-invariant and employed salient region-extraction, it could not handle conditional variations. In this work, we investigate whether CoHOG and SeqSLAM can be combined to complement each other. By combining these two powerful VPR techniques, our paper shows that it is indeed possible to achieve a fully handcrafted, training-less and light-weight VPR system, yielding state-of-the-art performance on both viewpoint- and conditionally-variant datasets. This blend of SeqSLAM and CoHOG is labelled as ConvSequential-SLAM. Fig. \ref{fig:diagram} shows the block-diagram of ConvSequential-SLAM.

        Formally, in this paper, by deriving motivation from the work of Milford \textit{et al.} \cite{milford2012seqslam} and Zaffar \textit{et al.} \cite{zaffar2020cohog}, we make the following main contributions:
        
        \begin{enumerate}
            \item We integrate convolutional scanning into SeqSLAM, achieving viewpoint invariance.
            \item We improve the conditional invariance of SeqSLAM by replacing contrast-enhanced pixel-matching with regional, block-normalised Histogram-of-Oriented-Gradients (HOG) descriptors.
            \item We introduce sequential matching into CoHOG to improve conditional invariance.
            \item We analyse the information-gain from consecutive query images to determine the minimum sequence length needed.
            \item Building upon the sequence length generated by analysing consecutive query images, we reuse the entropy computation for salient region extraction of CoHOG to formulate the most optimal dynamic sequence length, instead of a constant sequence length, as originally used in SeqSLAM. 
        \end{enumerate}

        The remainder of the paper is organised as follows: Section II presents an overview of the existing literature in VPR. Section III presents detailed information about ConvSequential-SLAM, while in Section IV we discuss the experimental setup. Section V presents the results and analysis obtained by evaluating the performance of our algorithm against other VPR techniques on public VPR datasets. The conclusion and future work are given in Section VI. 
       
    \section{Literature Review}
    \label{literature_review}
        A comprehensive review of the existing challenges and research in the field of VPR is presented in \cite{VPRsurvey}. \endgraf
        
        Local feature descriptors such as scale-invariant feature transform (SIFT) \cite{SIFT} and speeded-up robust features (SURF) \cite{SURF} make use of the most notable features in the image for extraction (keypoints), followed by description. These local descriptors have been widely used to perform VPR such as in \cite{se2002mobile}, \cite{andreasson2004topological}, \cite{stumm2013probabilistic}, \cite{kovsecka2005global}, \cite{murillo2007surf}. Different techniques have been combined together for the detection and description phase. In \cite{mei2009constant}, Mei \textit{et al.} have used FAST \cite{rosten2006machine} for detection, while the key-point extraction was achieved by SIFT descriptors. Similarly, in \cite{churchill2013experience}, the authors have used FAST for feature detection followed by BRIEF \cite{calonder2011brief} for feature description. FAB-MAP \cite{cummins2011appearance} is an appearance based place recognition system based on local feature descriptors integrated within a SLAM system. It represents visual places as words and uses SURF for feature detection. CAT-SLAM \cite{maddern2012cat}, extends the work of FAB-MAP by including odometry information. Center Surround Extremas (CenSurE) \cite{agrawal2008censure} introduces a suite of new feature detectors that outperforms the previously mentioned local feature descriptors, performing real-time detection and matching of image features. CenSurE has been used by FrameSLAM in \cite{konolige2008frameslam}. In reality, images may contain a large number of local features, and thus, to match these features together is not practically feasible. The Bag-of-Words model (BoW) \cite{sivic2003video}, \cite{fei2005bayesian} overcomes this issue by allocating similar visual features in corresponding bins. The resulting description of the environment using this technique is pose invariant, as spatial information in the descriptors is abstracted. BoW has been used for VPR tasks such as in \cite{angeli2008incremental}.  \endgraf
       A very popular global feature descriptor is Gist \cite{globalimagefeatures}, \cite{oliva2001modeling}, which uses Gabor filters in order to create a vector that will ultimately represent the content of an image. The work done in \cite{murillo2009experiments}, \cite{singh2010visual} and \cite{siagian2009biologically} shows some examples of Gist whole-image descriptor used in place recognition. As BRIEF \cite{calonder2011brief} holds comparable recognition accuracy with both SIFT and SURF but at reduced encoding time, it has been paired with Gist in \cite{sunderhauf2011brief}. The result is used as the front-end for a large scale SLAM system.  Badino \textit{et al.} \cite{badino2012real} proposed a variation of SURF, named Whole-Image SURF (WI-SURF), that integrates the accuracy resulting from metric methods together with the robustness of topological localisation, in order to perform visual localisation. The authors of SeqSLAM \cite{milford2012seqslam} decided to compare sequences of camera frames instead of single image comparison, thus achieving increased performance in VPR compared to traditional feature-based techniques, when the place is subject to drastic changes. The work of Pepperell \textit{et al.} \cite{pepperell2014all} on SMART extended SeqSLAM by incorporating into the calculations the varying speed of a vehicle. Histogram-of-Oriented-Gradients (HOG) \cite{dalal2005histograms} is a global descriptor which creates a histogram by calculating the gradient of all pixels present in the image. McManus \textit{et al.} used HOG for VPR in \cite{mcmanus2014scene} and Milford \textit{et al.} in \cite{milford2012seqslam}. \endgraf
        
        CNNs are known to be robust feature extractors and their performance on VPR related tasks showed 
        promising results, thus being extensively explored in the field of place recognition in challenging 
        environments. Because of their ability to learn generic features, researchers have extensively 
        explored the performance of CNNs on various visual tasks \cite{sharif2014cnn}, 
        \cite{oquab2014learning}. The authors of \cite{chen2014convolutional}, combined all 21 layers of the
        Overfeat network \cite{sermanet2013overfeat} trained on ImageNet 2012 dataset together with the 
        spatial and sequential filter of SeqSLAM. Chen \textit{et al.} \cite{chen2017deep} trained two 
        neural network architectures, namely HybridNet and AMOSNet whose performance in the field of place 
        recognition was assessed on the Specific PlaceEs Dataset (SPED). The initialization of the top 5 
        convolutional layers are based on the weights learnt from CaffeNet \cite{krizhevsky2012imagenet}, which is considered to drastically improve the features' robustness. The first neural network used in this work, 
        namely HybridNet, used the aforementioned initialised weights of CaffeNet, while AMOSNet was initialised with random weights. Arandjelović \textit{et al.} \cite{arandjelovic2016netvlad} introduced a 
        new layer based on a generalised Vector of Locally Aggregated Descriptors (VLAD) entitled NetVLAD, 
        that can be incorporated in any CNN architecture for VPR training. NetVLAD can be broken down into 
        two main stages as follows: the first stage is represented by a CNN which has the task of extracting the 
        features from images, while the second stage is represented by a layer that combines these features 
        in order to create an image descriptor. The authors of \cite{zaffar2019levelling} tested the 
        performance of NetVLAD on multiple datasets, including: Berlin Kudamm, Gardens Point and Nordland 
        datasets, showing its robust performance given various VPR scenarios. Until recently, the description 
        of places was achieved using whole-image description. The focus has changed to using Regions Of 
        Interest (ROI) which shows important improvement in performance when compared with the whole-image 
        description of places. In Cross-Region-BoW \cite{chen2017only}, the authors identified the ROI of 
        a query image using the CNN's layers acting as high-level feature extractors, then described these 
        regions using low-level convolutional layers. Khaliq \textit{et al.} \cite{khaliq2019holistic} 
        presents a lightweight VPR approach, based on ROI extraction combined with VLAD in order to achieve 
        state-of-the-art performance under severe viewpoint and conditional variations. CALC 
        \cite{merrill2018lightweight} trained a Convolutional Auto-Encoder to output illumination-invariant Histogram-of-Oriented-Gradients (HOG) descriptors, where instead of using the original version of the image, laterally shifted and distorted versions of the image are used as input to output the same HOG descriptor for all distorted inputs. This results in a very light-weight system robust to 
        variations in viewpoint and illumination. RatSLAM \cite{milford2008mapping}, is a biologically inspired neural network-based place recognition technique, which is based on the cell structure of a rat's hippocampus. The neural network used in this work is a  Continuous Attractor Network (CAN), its main task being to model the place cells. The authors of \cite{giovannangeli2006robust} employed the place cell model in their search of achieving vision-based navigation, without the need of a metric map. Even though RatSLAM was a success, the performance of the system suffers when place recognition is performed during day-to-night transitions. \endgraf
        Authors of CoHOG \cite{zaffar2020cohog} proposed a training-free technique based on the 
        Histogram-of-Oriented-Gradients (HOG) descriptor that is able to achieve state-of-the-art 
        performance in VPR. Using the information content (entropy) of each query image, together with the 
        region based convolutional matching, CoHOG is able to successfully perform VPR in challenging 
        environments. In this work, we combine CoHOG and SeqSLAM to explore the possibility of matching dynamic sequences of query and 
        reference frames under changing viewpoint and appearance conditions and report state-of-the-art performance in comparison to a number of contemporary VPR
        techniques on public VPR datasets.        
    
    \section{Methodology}
\begin{figure}
        \centering
        \includegraphics[width=1\linewidth]{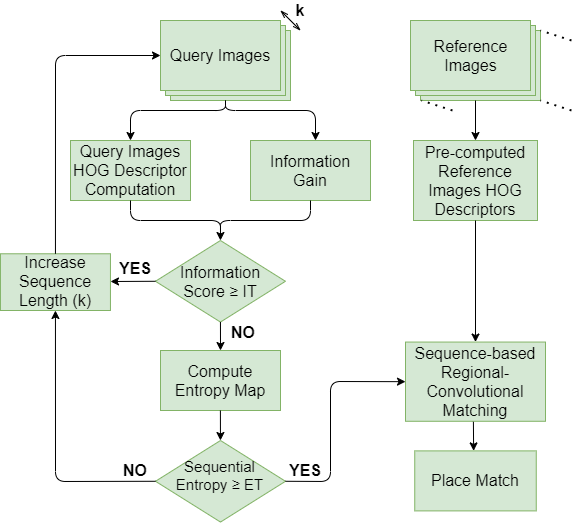}
        \caption{The block diagram of our framework is given here, which presents all the major components of the system.} 
        \label{fig:diagram}
        \end{figure}

        This section presents the methodology proposed in our work. The query images represent the visual data received from the camera, while the reference images represent the stored map of the environment in the form of RGB images. The block diagram showing each step of the ConvSequential-SLAM system is presented in Fig. \ref{fig:diagram}. We proceed to show the implementation of each one of these steps in the following subsections, namely Information Gain, Entropy Map and ROI Extraction, Regional HOG Computation, Regional Convolutional Matching, Creating the 1D Query list, Computing Information Content for a Dynamic Sequence of Images and Dynamic Sequence Matching.
        
        \subsection{Information Gain}
            The first major innovation in our work is the ability of our technique to determine the information-gain resulted from analysing consecutive query images. This allows a more robust understanding of the environment, while it also gives enough information about different textures and properties found in successive query images. This approach is used to determine the local change-point in consecutive query images, thus enabling a minimum sequence length (\textit{min\_k}) for each sequence of images to be determined (see subsection E). \endgraf 
            The information-gain is calculated as follows. Firstly, we compute the Histogram-of-Oriented-Gradients (see subsection C) of first query image that is part of a sequence. We then proceed to apply the same operation to the following query image, then comparing these two images together using regional convolutional matching (see subsection D). Finally, we compare the similarity score generated by the regional convolutional matching process with the Information Threshold (IT) to determine if the similarity between the two query images provides sufficient information gain. We then proceed to compare the first query image with the third and so on, until we find a representative sequence length. The information gain can be easily summarised as in equation (\ref{eq:infogain}) and (\ref{eq:infothreshold}) below:
            \medskip
            \begin{equation}\label{eq:infogain}
                Information\ Gain = 1 - Similarity\ Score
            \end{equation}
            
            \begin{equation}\label{eq:infothreshold}
                    Initial\_Seq\_Length = 
                    \left\{
                    \begin{array}{ll}
                        k + 1,&\mbox{if $Information\ Gain \geq IT$}.\\
                        entropy\ map,&\mbox{otherwise}.
                    \end{array}
                    \right.
                \end{equation}
            \smallskip
                

            In the above equation, $min\_k \leq k \leq max\_k\_info\_gain$, \textit{IT} represents the Information Threshold and it is in range [0,1], \textit{min\_k} is the minimum sequence length (set to 1) and \textit{max\_k\_info\_gain} is the maximum sequence length. The Initial Sequence Length (Initial\_Seq\_Length) in equation (\ref{eq:infothreshold}) represents the number of query images that are part of the query list generated by this approach. When the Information Gain module provides its best sequence length  (e.g. $Information\ Gain < IT$), we proceed to calculate the sequential entropy (see subsection F) for that sequence of query images and determine whether this has the optimal length.
            
        \subsection{Entropy Map and ROI Extraction}
            The second step in the ConvSequential-SLAM framework is to create the entropy map representing the salient regions in each query image. This is similar to \cite{zaffar2020cohog}, where the entropy map creation is based on estimating the local pixel intensity variation within the grayscale image and computing the base-2 logarithm of the histogram of pixel intensity values within each local region. This entropy map is represented by a matrix of size \textit{W1*H1}, the elements of which are values in the range \{0-8\}, due to the pixel intensities in the range of $2^0$ to $2^8-1$. The dimensions \textit{W1*H1} represent the fixed size dimensions of the input image. The following matrix represents the entropy map for a query image:
            
            \[
            Entropy = 
            \begin{bmatrix}
                e_{11} & e_{12} & e_{13} & \dots  & e_{1W1} \\
                e_{21} & e_{22} & e_{23} & \dots  & e_{2W1} \\
                \vdots & \vdots & \vdots & \ddots & \vdots \\
                e_{H1W1} & e_{H1W1} & e_{H1W1} & \dots  & e_{H1W1}
            \end{bmatrix}
            \]
            \begin{center}
                Where $e_{ij} \in \{0-8\}$.
            \end{center}
            
            Using the entropy map of an image, we extract Regions-of-Interest (ROIs) by computing the average entropy of a region of size \textit{W2*H2}. If this entropy is above a threshold $ET$, it reflects that a region in informative and is selected as an ROI. The total number of regions (non-overlapping) in an image is $N=W1/W2 \times H1/H2$ and the total number of ROIs is $G$ which can vary from one query image to another. Further details about the effect of $ET$ and entropy map creation have been discussed at length in \cite{zaffar2020cohog} and not provided here to avoid redundancy.
            
            Moreover, in order to get a single entropy value for the entire image, we sum all the elements of the entropy matrix and divide them by \textit{W1*H1*8} to get the re-scaled value. This is useful for the computation of sequential entropy of a sequence of query images to determine the dynamic sequence length (see subsection F).
            
        \subsection{Regional HOG Computation}
            The process of regional HOG computation takes place as follows. In the first instance, we compute a gradient map of a grayscale image of size \textit{W1*H1}. Following this, a histogram of oriented-gradients is computed for all \textit{N} regions of the image, with each region having the size of \textit{W2*H2}. Furthermore, each histogram of every region has \textit{L} bins, where each bin is labelled with equally spaced gradient angles between 0-180 degrees. Lastly, we use L2-normalisation to achieve illumination invariance. This is done at a block level of size \textit{(W2*2)*(H2*2)}.
            
        \subsection{Regional Convolutional Matching} 
            Following the regional HOG computation, we proceed to regional convolutional matching, given each query image is represented as \textit{N} regions, each being described by a HOG-descriptor of depth 4*\textit{L}. Using the information from the Region of Interest (ROI) evaluation, these \textit{N} regions are reduced to \textit{G} salient regions. By doing so, the query image HOG-descriptor can be represented as a 2D matrix of dimensions [\textit{G}, 4*\textit{L}]. The reference image has \textit{N} regions with the descriptor size of 4*\textit{L}, therefore its resulting matrix has the dimensions of [\textit{N}, 4*\textit{L}]. We then proceed to multiply the query and reference matrices, and the result is a matrix of dimensions [\textit{G}, \textit{N}]. Each row of this matrix represents a salient region of a query image, while each column represents the cosine-matching scores for that region with all the \textit{N} regions of a reference image. Max-pooling is used across the rows of the aforementioned matrix in order to determine the best matched regions between the query and reference images. The final score is computed as the arithmetic mean of matching scores of all \textit{G} regions and is in the range of 0 - 1, such that the higher the score, the higher the similarity between the two images. Finally, the reference image that has the highest score is chosen to be the best match for a given query image.
            
        \subsection{Creating the Query Images Sequence}
            Query images are added into a 1D list in a sequential manner, such that the length of this list is dependent on the sequential entropy (explained in subsection F). Even if the sequential entropy's value for the first \textit{k} images is higher than the Entropy Threshold (\textit{ET}), where $0 \leq ET \leq 1$, the minimum sequence length will be determined using the information-gain resulted from analysing consecutive query images (see subsection A), in such a manner that we will not end up with non-optimal sequence lengths, that will ultimately result in poor performance (see equation (\ref{eq:sequential})). 
            The 1D query list containing a sequence of query images is represented as:
            \bigskip
            \[ Sequential\ Query\ List = 
            \begin{bmatrix}
                q_{1} & q_{2} & q_{3} & \dots  & q_{k}
            \end{bmatrix}
            \]
          \medskip
            \par
            In the above equation $q_{1}$ is the first query image, $q_{k}$ is the last query image, and \textit{k} is the total number of images that are part of a sequence. \endgraf
            As previously mentioned, the length of this list will constantly change, but all the images will be in a sequential order, starting from the first image to the \textit{k-th} image. When computing the second sequence of query images, we start with the second image ($q_{2}$) and so on. It is important to note that for any \textit{N} images read, the number of query images sequence lists created will be \textit{N} - \textit{k} + \textit{1}, where \textit{k} will contain the length of the last list created. That is, for any \textit{N} query images, the algorithm will only match the first \textit{N} - \textit{k} + \textit{1} images. 
        
        \subsection{Entropy-based Dynamic Query Images Sequence}
        
            The second key innovation is incorporating the ability of our technique to reuse entropy as
            measure of the overall information content found in a sequence of query images, to decide the most optimal sequence length of each query list. Building upon the sequence length generated by analysing consecutive query images (see subsection A), we use the entropy to maximize the efficiency of this length. To achieve this, our technique first looks at the information content (entropy score) of the query sequence list generated by the information-gain approach (first \textit{min\_k\_info\_gain} query images). If the information content within this sequence of images is less than a threshold ($ET$), we increase the sequence length by a constant step, then recompute the information content for this new increased sequence of images. If the information content ($Seq\_Entropy$) for this increased sequence of images reaches a reasonable value ($ET$), the corresponding length of the query images sequence is used, otherwise we keep increasing it (up till maximum sequence length) to find a suitable sequence length. $Seq\_Entropy$ represents the arithmetic mean of the entropy scores of the query images within the sequence. The entire 
            iterative process is summarised in equation (\ref{eq:sequential}) below:
           \bigskip
                \begin{equation}\label{eq:sequential}
                    Seq\_Length =
                    \left\{
                    \begin{array}{ll}
                        min\_k\_info\_gain,&\mbox{if $Seq\_Entropy \geq ET$}.\\
                        k + 1,&\mbox{otherwise}.
                    \end{array}
                    \right.
                \end{equation}
            \medskip
            
                
        
        In the above equation $min\_k\_info\_gain \leq k \leq max\_k$, \textit{ET} represents the Entropy Threshold, \textit{min\_k\_info\_gain} is the minimum sequence length (generated in subsection A) and \textit{max\_k} is the maximum sequence length. The Sequence Length (\textit{Seq\_Length}) in equation (\ref{eq:sequential}) represents the number of images that are part of the query list at a given time, thus being dependent on the value of \textit{k}. In the same equation, the Sequential Entropy (\textit{Seq\_Entropy}) refers to the average entropy value (see subsection B) of \textit{k} images that are part of this query list.
         
        \subsection{Dynamic Sequence Matching}
            This subsection presents how we achieve the matching between dynamic sequence length of images. As discussed in subsection F, our technique creates a dynamic list of query images, i.e, the length of the query sequence list will vary for different sets of query images. During the matching phase, we create a sequential 1D reference list of the same length as the sequential query list. These sequential 1D reference lists are created for all the images in the reference map. Because the size of our reference list is dependent on the sequential query list's length, this simplifies the matching of the query and reference image sequences. The algorithm that retrieves a correct match for a sequence of query images given a reference map can be found in Algorithm 1. The function \textit{Sequence\_Matching\_Func} in Algorithm 1 takes \textit{k} corresponding pairs (1-to-1 matching) from the query image sequence ($Q\_Seq$) and reference image sequence ($R\_Seq$) and matches them using Regional Convolutional Matching, as explained in subsection D. The matching score of the query and reference sequences is the arithmetic mean of the matching scores of the pairs within these sequences. This function returns the matching score of the query image sequence and the reference image sequence. Given all the reference images and their corresponding sequences from the reference map, the sequence with the highest matching score is selected as the best match.
                    
    \section{Experimental Setup}
        
        To evaluate the performance of ConvSequential-SLAM in VPR related scenarios, we have
        used 4 public VPR datasets that pose difficulties for place matching techniques, as
        discussed in Section I. The first dataset used for testing the performance of our
        algorithm is the Gardens Point dataset \cite{sunderhauf2015performance}, containing images 
        taken from different angles (viewpoint variation). This dataset consists of a total
        of 600 images, a third of which are query images (day images) and the other two-thirds representing reference images (split into day and night images). In this 
        work, we have tested ConvSequential-SLAM on Gardens Point (day-to-day) and Gardens Point (day-to-night) datasets. Moving to seasonal variation, Nordland dataset \cite{nordland2013dataset}, captures the drastic visual changes of a place in different seasons (spring, summer, autumn and winter). Because the most important difference between seasons can be seen in summer and winter, we tested our
        algorithm on the summer-to-winter traverses of Nordland dataset. The fourth dataset used is Campus
        Loop dataset, which contains 100 query and 100 reference images. This dataset poses challenge to any VPR system due to the high amount of viewpoint variation, seasonal variation and also the presence of statically-occluded frames. Apart from using these 4 datasets to show the performance of our technique, we also use the Alderlay (night-to-day) dataset solely to show the variation in sequence length due to sequential entropy. This dataset consists of 201 consecutive query images (night images) and 201 consecutive reference images (day images). \endgraf
        \begin{algorithm}
            \SetAlgoLined
                \textit{Given:} Query Images Sequence (Q\_Seq) \\
                \textit{Given:} Reference\_Images\_List (R\_List) \\
                ref\_matching\_scores = $[\ ]$ \\
                iterator = 0 \\
                k = Length (Q\_Seq) \\
                \While{itr + k $\leq$ Length(R\_List)}{
                    Sequential\_Reference\_List = R\_Seq = $[\ ]$ \\
                    \For{ref\_itr \textbf{in range}(itr, itr + k)}{
                    APPEND R\_List[ref\_itr] to R\_Seq
                    }
                    match\_score = Sequence\_Matching\_Func (Q\_Seq, R\_Seq) \\
                    ADD match\_score to ref\_matching\_scores \\
                    iterator = iterator + 1
                }
                Best Match = Max (ref\_matching\_scores)
            \caption{Matching Query and Reference Sequences}
        \end{algorithm}
        In this work, we have used $ET=0.5$, $IT=0.9$, $min\_k=1$, $max\_k\_info\_gain=15$ and $max\_k=25$ for ConvSequential-SLAM and an ablation study of these is provided later in this paper. We compare the performance of ConvSequential-SLAM with other VPR techniques 
        reviewed by the authors of \cite{zaffar2019levelling}. The
        authors of \cite{VPRsurvey} suggested that Precision-Recall curves are a key evaluation metric for VPR techniques. Therefore, an ideal system would achieve 100$\%$ precision at 100$\%$ recall. The authors of \cite{zaffar2020vpr} used 
        the Area-under-the-Precision-Recall-Curve (AUC) to compare the performance of 
        VPR techniques which has therefore been adopted in our work as well. AUC-PR is computed by plotting the Precision-Recall curve at different confidence thresholds. Precision and Recall are determined using equation (\ref{eq:precision}) and (\ref{eq:recall}) respectively:
        \medskip
        \begin{equation}\label{eq:precision}
            Precision = \frac{True\ Positives}{True\ Positives + False\ Positives}
        \end{equation}    
        
        \begin{equation}\label{eq:recall}  
            Recall = \frac{True\ Positives}{True\ Positives + False\ Negatives}
        \end{equation}
        \smallskip

        In the above equations, the \textit{True Positives} are the correctly retrieved matches, \textit{False Positives} are the incorrectly retrieved matches and \textit{False Negatives} are the incorrectly discarded matches. \endgraf
        In \cite{zaffar2020vpr}, the authors highlighted that an ideal AUC score of 1 is achievable even if the system contains false-positives and therefore suggested that the accuracy \textit{A} (refer to equation (\ref{eq:accuracy})) of a system should also be provided. Moreover, AUC-PR is only focusing on the matching performance of a given VPR system, thus not incorporating the computational intensity of that technique. This is essential in real-world scenarios, where for resource-constrained platforms, the matching performance has to be directly related with the computational intensity. For this reason, the authors of \cite{zaffar2019state}, \cite{zaffar2019levelling}, \cite{khaliq2019holistic} and \cite{merrill2018lightweight} determined the feature encoding time ($t_e$) to be an important performance indicator. In \cite{zaffar2020cohog}, the Performance-per-Compute-Unit (PCU) is defined by combining P$_{R100}$ with $t_e$ as in equation (\ref{eq:PCU}):
    \medskip
            \begin{equation}\label{eq:accuracy}
               A = \frac{Total\ Correctly\ Matched\ Query\ Images\ in \ Database}{Total\ No.\ of\ Query\ Images \ in \ Database}
            \end{equation}
            
        \begin{equation}\label{eq:PCU}         
            PCU =  P_{R100}\ \times \ \log\bigg(\frac{t_{e\_max}}{t_e} + 9\bigg)
        \end{equation}

        As it can be seen in the above equation, higher precision will always lead to a 
        higher PCU value. The maximum feature encoding time ($t_{e\_max}$) is chosen to 
        represent the most resource intensive VPR technique. In our case, this is 
        NetVLAD, with the feature encoding time of $t_{e\_max}$ = 0.77. Fig. 5 shows the
        feature encoding time ($t_e$) for each VPR technique tested. The scalar 9
        is added in equation (\ref{eq:PCU}) so that the VPR technique that has $t_e$ = $t_{e\_max}$ does not have a PCU = 0, but instead that it provides a more interpretable range.
        
    \section{Results and Analysis}
        This section is aimed at providing a comparison of ConvSequential-SLAM with other state-of-the-art VPR techniques on all datasets mentioned in Section IV. We discuss these results from a place matching performance point, in terms of accuracy, AUC-PR and PCU. We also present the performance of ConvSequential-SLAM for various sequence lengths and show how the sequence length varies between one dataset and another. Finally, we show some samples of correctly and incorrectly retrieved query and reference images by our technique for a qualitative insight.
            
        \subsection{Accuracy}
          
            \begin{figure*}[t]
            \centering
            \includegraphics[width=\textwidth]{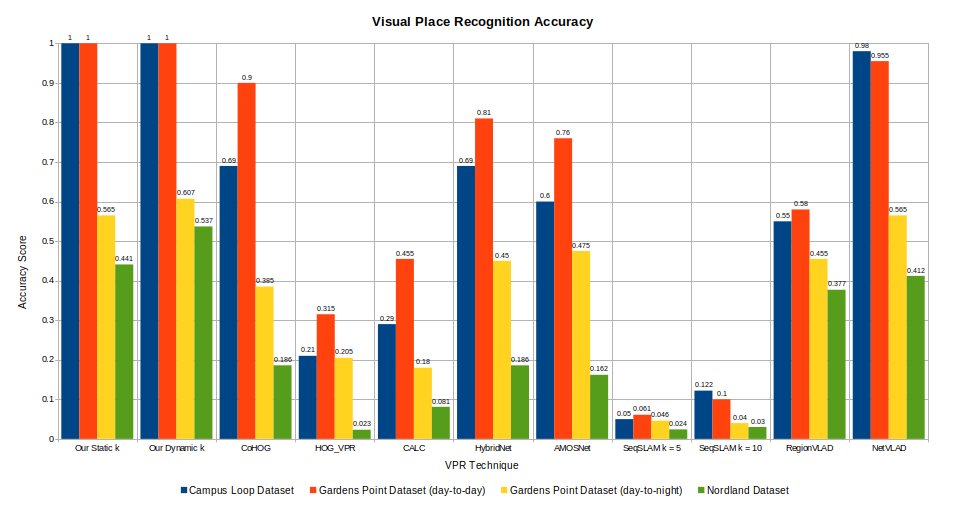}
            \caption{The Accuracy of ConvSequential-SLAM (using a fixed sequence length of 10 images as well as a dynamic length determined by the system itself) is compared against the accuracy of other state-of-the-art VPR techniques (including CoHOG, HOG VPR, CALC, HybridNet, AMOSNet, SeqSLAM, RegionVLAD and NetVLAD) on Campus Loop (day-to-day), Gardens Point (day-to-day), Gardens Point (day-to-night) and Nordland (summer-to-winter) Datasets (mentioned in section IV).}
            \label{accuracy_all}
            \end{figure*}
        
            This subsection presents the accuracy results of ConvSequential-SLAM against the performance of other VPR techniques. Fig. \ref{accuracy_all} shows the computed values of accuracy for all techniques, on all 4 datasets. As all the datasets tested contain consecutive images, there is a high possibility that each image is similar to the ones located in its immediate proximity. Therefore, if for any query image, the reference image found to be the best match is in the range $\pm 2$, we consider it as a correct match, except for the Nordland dataset where we use the $\pm 1$ range. \endgraf
          
            ConvSequential-SLAM achieves state-of-the-art accuracy on all datasets utilised in this work. It is especially important to see this comparison with CoHOG, where due to the dynamic use of sequential information (instead of single frames), our technique performs much better. The results suggest that our blend of CoHOG and SeqSLAM performs significantly better than either of these in their individual capacity. Our approach also achieves state-of-the-art performance on the highly conditionally-variant Gardens Point (day-to-night) and Nordland datasets, followed by state-of-the-art deep learning-based techniques like NetVLAD, HybridNet and AMOSNet.  
            
        \subsection{Area-Under-the-Precision-Recall-Curve}
            This subsection reports the performance of ConvSequential-SLAM in terms of AUC-PR on all datasets in Table I. Our technique achieves state-of-the-art AUC-PR performance on both Campus Loop Dataset, as well as Gardens Point (day-to-day) and Gardens Point (day-to-night) datasets. When compared to the existing state-of-the-art NetVLAD, ConvSequential-SLAM achieves better performance on all datasets tested, except on Nordland dataset, as reported in Table I. We can see a light boost in performance between our algorithm using a fixed sequence length of 10 images and a dynamic sequence length respectively. In Fig. \ref{pr_curves}, we present the Precision-Recall curves of all the VPR techniques tested in our work on all 4 datasets introduced in Section IV. 
             \begin{table}
            \caption{The AUC-PR of VPR techniques on the 4 datasets.} 
            \includegraphics[width=252pt]{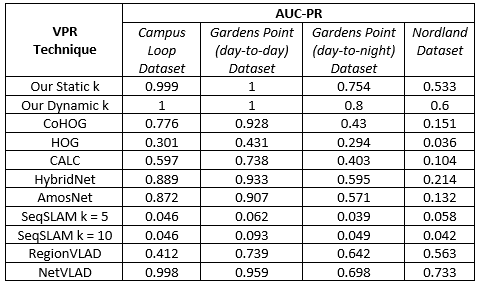}
            \end{table} 
        \subsection{Performance-Per-Compute-Unit (PCU)}
\begin{figure}
        \includegraphics[width=252pt]{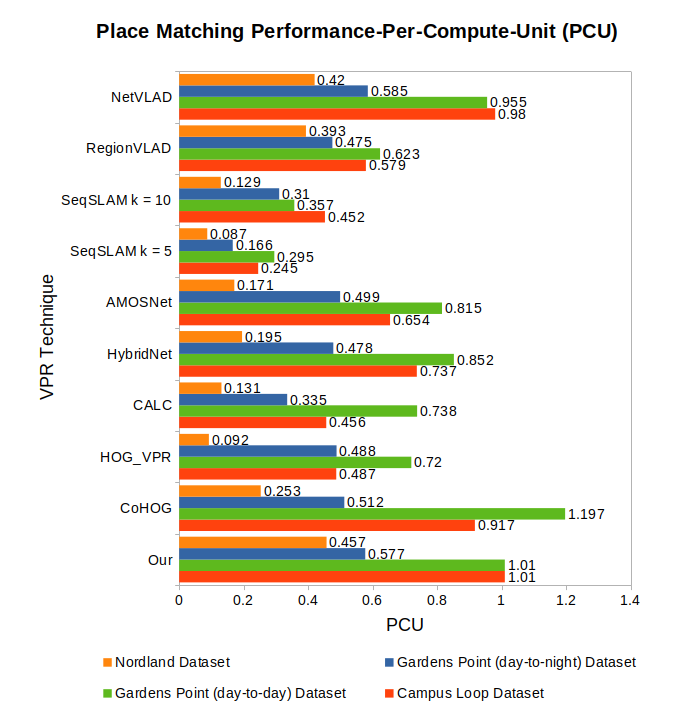}
        \caption{The PCU of ConvSequential-SLAM is compared with the PCU of other contemporary VPR 
        techniques on all mentioned datasets.} 
        \label{PCU}
    \end{figure}

\begin{figure}
            \centering
            \includegraphics[width=252pt]{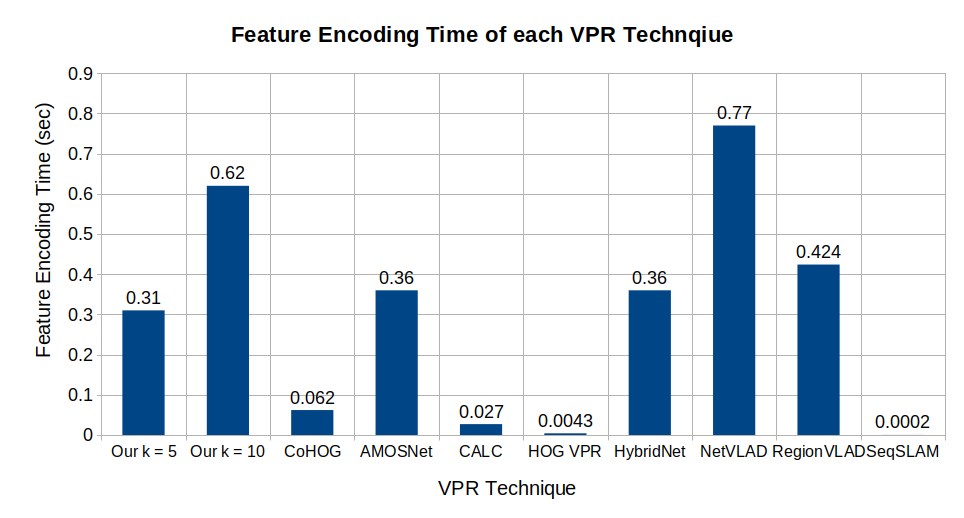}
            \caption{The feature encoding time of each VPR technique used in this work is 
                shown in this graph. \textit{Our k} = 5 and \textit{k} = 10 represent the feature encoding time 
            of ConvSequential-SLAM using fixed sequences of 5 and 10 images respectively.} 
            \label{encodingtimes}
    \end{figure}

\begin{figure*}
            \centering
            \begin{tabular}{ c c }
                 \includegraphics[width=252pt]{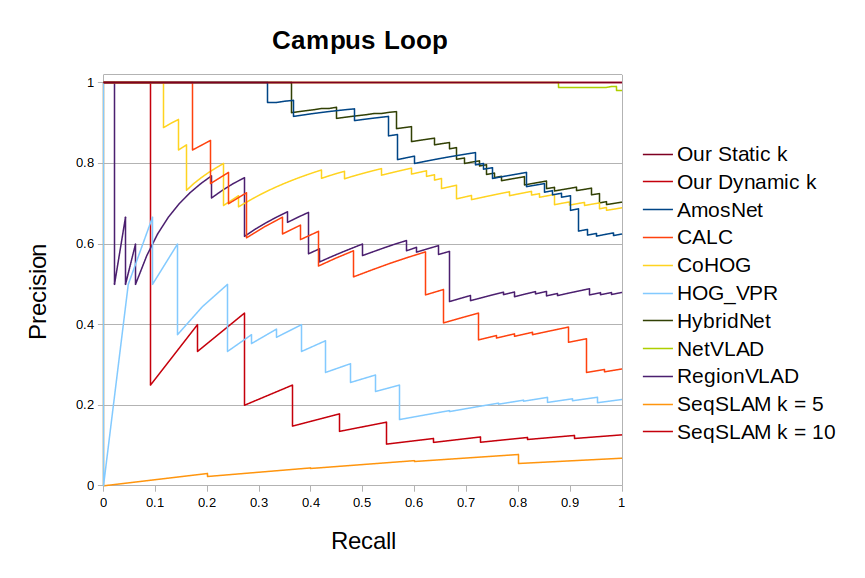} &  
                 \includegraphics[width=252pt]{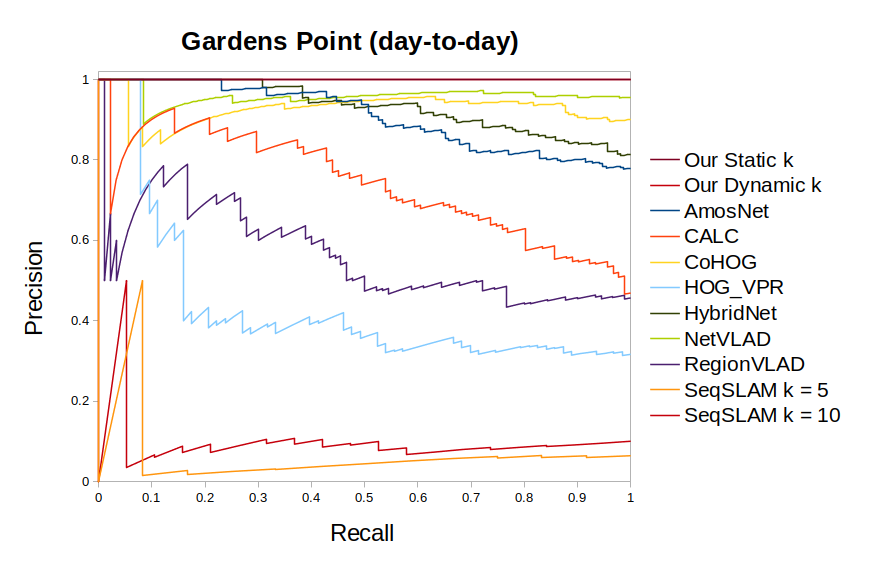} \\ 
                 \includegraphics[width=252pt]{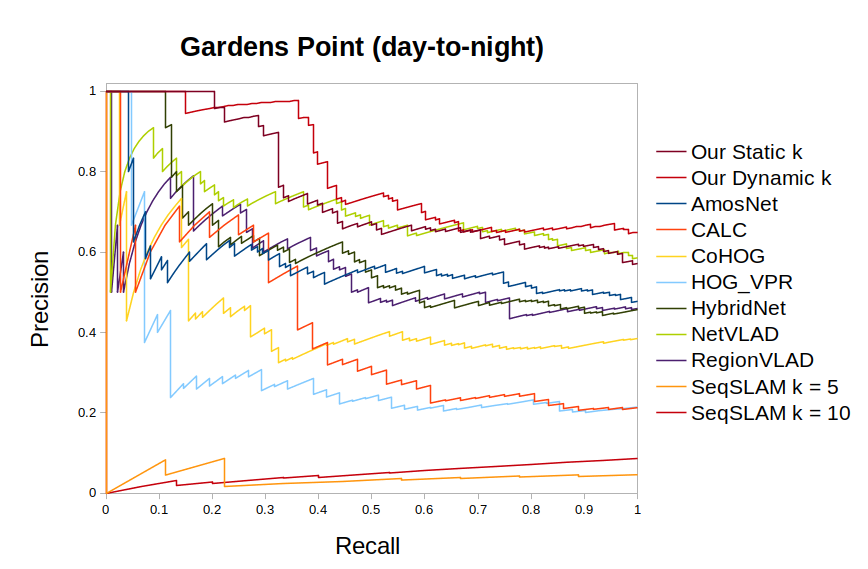}  &
                 \includegraphics[width=252pt]{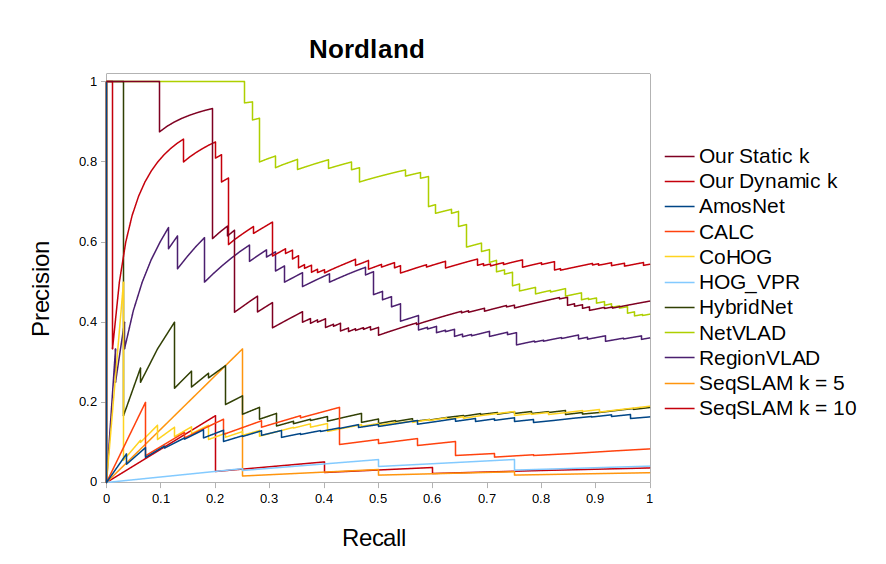}  
                      
            \end{tabular}
            \caption{The Precision-Recall Curves for all VPR techniques on each of the 4 datasets used in our work are enclosed here. }
            \label{pr_curves}
            \end{figure*}

            Fig. \ref{PCU} presents the PCU of ConvSequential-SLAM using a fixed sequence length of 10 images. Because we match sequences of images instead of single frames, the feature encoding time will also be increased with each image that is part of that sequence, as shown in Fig. \ref{encodingtimes}. The feature encoding time for dynamic k will vary between the lowest value (that is for a minimum sequence length determined by the information-gain) and the maximum value for a sequence length of 25 images. \endgraf
            In this subsection, we use the average sequence length computed by our methodology within the dataset for encoding time computation. Because the precision of ConvSequential-SLAM has increased when compared to CoHOG, the PCU value has also increased, even though the encoding time has been raised by \textit{k} folds.

        \subsection{Variation in sequence length}
\begin{figure}
            \centering
            \includegraphics[width=252pt]{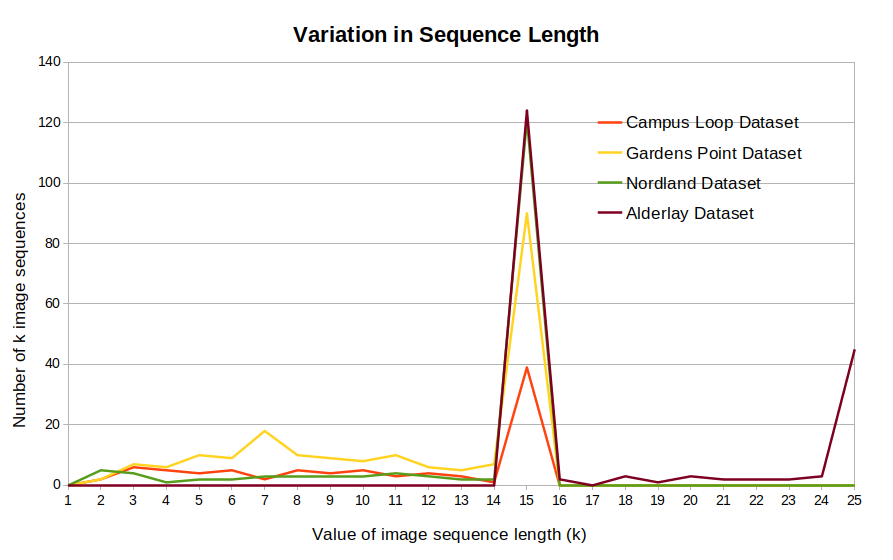}
            \caption{The variation in sequence length of ConvSequential-SLAM on all four datasets is shown here.} 
            \label{variationinseqlength}
            \end{figure}

            Fig. \ref{variationinseqlength} presents the different sequence lengths (\textit{k}) that can be achieved when the program was run on all datasets. It is important to note that by varying both the Entropy Threshold (\textit{ET}) and Information Threshold (\textit{IT}) we can achieve lower or higher sequence lengths. Also, it is worth noting that we use day images as query images for both Gardens Point (day-to-day) and Gardens Point (day-to-night) datasets, so we only include one instance of the dataset in Fig. \ref{variationinseqlength} in order to avoid redundancy. \endgraf
            In addition to the datasets mentioned above, we also use the Alderlay (night-to-day) dataset to show how the sequence length is modifying because of entropy. However, all VPR techniques poorly perform on this dataset, because the query images (night images) provide little to no information about the environment. This is due to the poor lighting condition in the environment as well as the presence of rain, which increase the difficulty in place matching.  \endgraf
            Because on Campus Loop, Gardens Point and Nordland datasets the entropy across each dataset is too high, the sequence length would not have increased in most cases, therefore we would end up with non-optimal sequence lengths. By using information-gain resulted from analysing consecutive query images, we are able to increase the minimum sequence length even though the salient information found in any given query image is above the threshold set (e.g. \textit{ET} $\geq$ 0.5). However, in contrast with the previously mentioned datasets where the sequence length would not increase due to high information content in query images, on the Alderlay dataset query images (night images) do not contain salient information due to poor illumination, therefore the sequence length will increase up to the maximum sequence length of 25 as shown in Fig. \ref{variationinseqlength}. \endgraf
            Using entropy is particularly helpful in scenarios where the query frames do not provide much information (as mentioned above), thus increasing the sequence length allows better chances of finding the correct reference image for any given query image. On the other hand, in cases where the information content of multiple sequences of query images is too high in a dataset (such as Campus Loop, Gardens Point and Nordland datasets), non-optimal sequence lengths will be achieved if entropy alone is used. Therefore, information gain is used to establish the lower bounds of the sequence length needed in order to achieve the most optimal results. 
        
        \subsection{Ablation Study}
        \label{ablation_study}
\begin{figure}
            \centering
            \includegraphics[width=252pt]{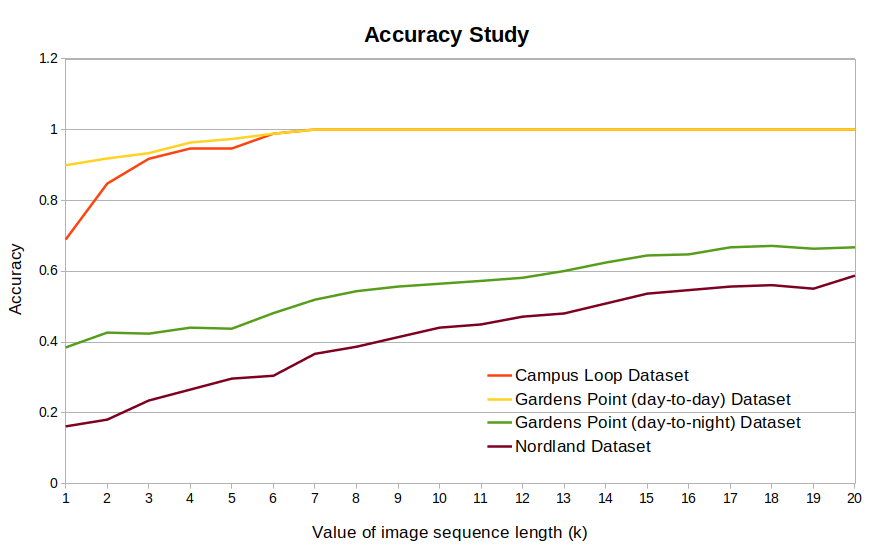}
            \caption{The ablation study showing the accuracy of ConvSequential-SLAM for different $k$ values ($1 \leq k \leq 20$).} 
            \label{ablationaccuracy}
            \end{figure}

\begin{figure} 
            \centering
            \includegraphics[width=252pt]{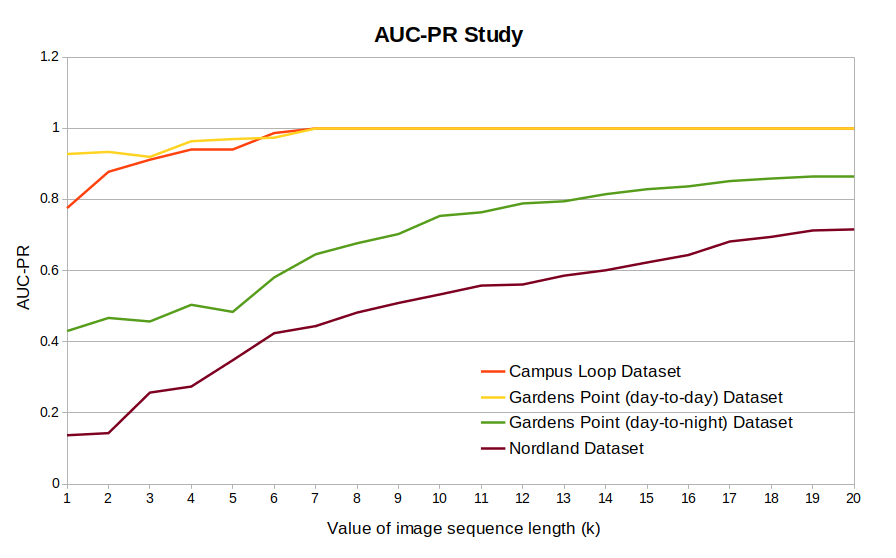}
            \caption{The ablation study showing the AUC-PR of ConvSequential-SLAM for different $k$ values ($1 \leq k \leq 20$).} 
            \label{ablationaucpr}
            \end{figure}

            Fig. \ref{ablationaccuracy} and Fig. \ref{ablationaucpr} presents the performance of our approach in terms of Accuracy and AUC-PR values, when using dynamic length of images. The program was tested with fixed k lengths between 1 and 20 images respectively. It is important to note that for k = 1, ConvSequential-SLAM's performance does not differ from CoHOG and has the exact same results. Increasing the value of k leads to an increase in both Accuracy and AUC-PR performance.
        
           
           
        \subsection{Exemplar Matches}
\begin{figure} 
         \centering
            \includegraphics[width=252pt]{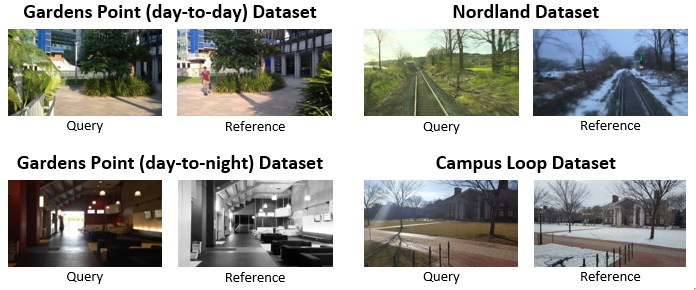}
            \caption{Correctly matched query and reference frames as taken from different datasets. } 
            \label{correctlymatchedimages}
        \end{figure}

\begin{figure} [h]
         \centering
            \includegraphics[width=252pt]{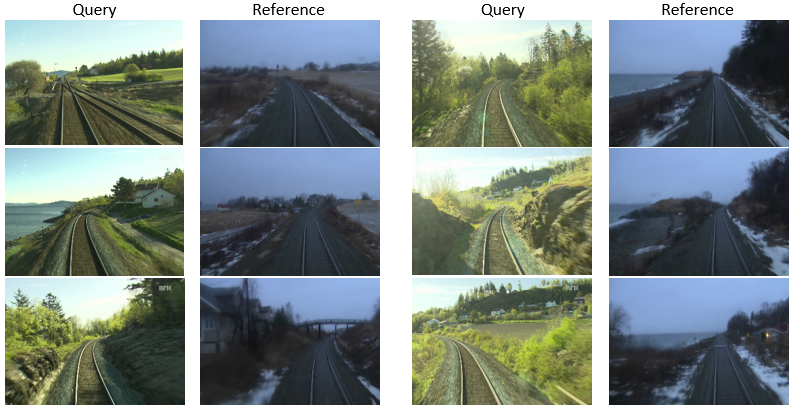}
            \caption{Sequence of incorrectly matched query and reference frames, taken from the Nordland dataset. } 
            \label{incorrectlymatchedimages}
        \end{figure}

            Fig. \ref{correctlymatchedimages} shows some correctly matched query and reference images, taken from each dataset. As mentioned in section IV, these datasets contain viewpoint, seasonal and illumination variations as well as confusing frames. We show that our approach successfully performs VPR in these challenging conditions. \endgraf
            Having shown the ability of our system to successfully perform VPR in challenging scenarios, we now proceed to show some failure examples, in which our algorithm incorrectly assigns the correct query and reference images. Fig. \ref{incorrectlymatchedimages} shows a sequence of incorrectly matched query and reference frames in the Nordland dataset. Even though our technique successfully performs VPR under drastic viewpoint and illumination variations, the failure in matching is especially due to the presence of confusing features coming from trees and vegetation that can be found in most images throughout the Nordland dataset, increasing the difficulty in place matching. Having said that, these kinds of challenges have yet to be overcome by VPR techniques in order to achieve a fully functional system that is capable of successfully performing VPR in all scenarios.
        
    \section{Conclusion and Future Directions}
        In this paper, we present ConvSequential-SLAM, a system that successfully performs visual place recognition in challenging environments, with zero training requirements. Firstly, we analyse consecutive query frames (i.e. information gain) to determine the minimum sequence length needed. Secondly, we optimize this minimum sequence length using the entropy as a measure of overall content in a sequence of query images. We test the performance of ConvSequential-SLAM on public VPR datasets that contain both viewpoint and appearance variations. The results show that state-of-the-art performance is achieved by our technique (based on various performance metrics). While we are able to avoid feature-less, perceptually-aliased patches in images, our technique does not cater to dynamic objects and confusing features coming from trees, vegetation etc. in outdoor environments. Such dynamic and confusing features can be challenging for VPR as proposed in \cite{zaffar2018memorable} and identify possible future directions for improving our work.
        
    {
    \small
    \bibliographystyle{ieeetr}
    \bibliography{root}
    }

\end{document}